\documentclass[journal]{IEEEtran}

\usepackage{graphicx}
\usepackage{multirow}
\usepackage{multicol}
\usepackage{cite}
\usepackage{hyperref}
\usepackage{color}
\usepackage{amssymb}
\usepackage{url}
\usepackage{pdfpages}

\ifCLASSINFOpdf
\else
\fi

\begin{document}
	
	\bstctlcite{IEEEexample:BSTcontrol}
	
	\title{A Deep Attentive Convolutional Neural Network for Automatic Cortical Plate Segmentation in Fetal MRI}

	\author{
		Haoran Dou, Davood Karimi, Caitlin K. Rollins, Cynthia M. Ortinau, Lana Vasung, Clemente Velasco-Annis, Abdelhakim Ouaalam, Xin Yang, Dong Ni, Ali Gholipour,~\IEEEmembership{Senior Member,~IEEE}
		\thanks{
			© 2020 IEEE.  Personal use of this material is permitted.  Permission from IEEE must be obtained for all other uses, in any current or future media, including reprinting/republishing this material for advertising or promotional purposes, creating new collective works, for resale or redistribution to servers or lists, or reuse of any copyrighted component of this work in other works.
			
			This work was supported in part by the National Institutes of Health (NIH) under Grant R01 EB018988, Grant R01 NS106030, Grant K23 NS101120, and Grant K23 HL141602;	in part by the Technological Innovations in Neuroscience Award from the
			McKnight Foundation; in part by awards from the Mend A Heart Foundation; in part by the Pediatric Heart Network; in part by the American Academy of Neurology; in part by the Brain and Behaviour Research Foundation; in part by the National Key Research and Development Program of China under Grant 2019YFC0118300; and in part by the Shenzhen Peacock Plan under Grant KQTD2016053112051497 and Grant KQJSCX20180328095606003. The work of Lana Vasung was supported by the Ralph Schlager fellowship at Harvard University. 
			
			H. Dou is with the National-Regional Key Technology Engineering Laboratory for Medical Ultrasound, Guangdong Key Laboratory for Biomedical Measurements and Ultrasound Imaging, School of Biomedical Engineering, Health Science Center, Shenzhen University, Shenzhen, China; and with the Medical UltraSound Image Computing (MUSIC) Lab, Shenzhen University, China; and also with the Computational Radiology Laboratory in the Radiology Department at Boston Children's Hospital, Harvard Medical School, Boston, Massachusetts, USA (email:douhaoran2017@email.szu.edu.cn ).  
			
			D. Karimi, and A. Gholipour are with the Computational Radiology Laboratory (CRL) in the Department of Radiology at Boston Children's Hospital, and Harvard Medical School, Boston, Massachusetts, USA (email: davood.karimi@childrens.harvard.edu, ali.gholipour@childrens.harvard.edu). 
			
			C.K. Rollins is with the Department of Neurology at Boston Children's Hospital, and Harvard Medical School, Boston, Massachusetts, USA.
			
			C.M. Ortinau is with the Department of Pediatrics in Washington University School of Medicine in St. Louis, Missouri, USA.
			
			L. Vasung is with the Department of Pediatrics at Boston Children's Hospital, and Harvard Medical School, Boston, Massachusetts, USA.
			
			C. Velasco-Annis, and A. Ouaalam are with the CRL in the Department of Radiology at Boston Children's Hospital, Boston, Massachusetts, USA.
			
			X. Yang, and D. Ni are with the National-Regional Key Technology Engineering Laboratory for Medical Ultrasound, Guangdong Key Laboratory for Biomedical Measurements and Ultrasound Imaging, School of Biomedical Engineering, Health Science Center, Shenzhen University, Shenzhen, China, and also with the MUSIC Lab, Shenzhen University, China (email: nidong@szu.edu.cn)
			
			Corresponding authors: Dong Ni and Ali Gholipour.
			
			The content is solely the responsibility of the authors and does not necessarily represent the official views of the NIH or the foundations.
			
			The code for the techniques presented in this study can be found at: \url{https://github.com/bchimagine}, or \url{https://github.com/wulalago/FetalCPSeg}}
	}
	
	
	\maketitle
	
	\begin{abstract}
		Fetal cortical plate segmentation is essential in quantitative analysis of fetal brain maturation and cortical folding. Manual segmentation of the cortical plate, or manual refinement of automatic segmentations is tedious and time-consuming. Automatic segmentation of the cortical plate, on the other hand, is challenged by the relatively low resolution of the reconstructed fetal brain MRI scans compared to the thin structure of the cortical plate, partial voluming, and the wide range of variations in the morphology of the cortical plate as the brain matures during gestation. To reduce the burden of manual refinement of segmentations, we have developed a new and powerful deep learning segmentation method. Our method exploits new deep attentive modules with mixed kernel convolutions within a fully convolutional neural network architecture that utilizes deep supervision and residual connections. We evaluated our method quantitatively based on several performance measures and expert evaluations. Results show that our method outperforms several state-of-the-art deep models for segmentation, as well as a state-of-the-art multi-atlas segmentation technique. We achieved average Dice similarity coefficient of 0.87, average Hausdorff distance of 0.96 mm, and average symmetric surface difference of 0.28 mm on reconstructed fetal brain MRI scans of fetuses scanned in the gestational age range of 16 to 39 weeks (28.6$\pm$5.3). With a computation time of less than 1 minute per fetal brain, our method can facilitate and accelerate large-scale studies on normal and altered fetal brain cortical maturation and folding.
	\end{abstract}
	
	\
	\IEEEpeerreviewmaketitle
	
	\section{Introduction}
	
	\IEEEPARstart{F}{etal} magnetic resonance imaging (MRI) has been established as a reliable method for quantitative evaluation of cortical development in the fetus, and for the diagnosis and analysis of congenital neurological disorders~\cite{coakley2004fetal}. It is a safe technique that provides much better soft tissue contrast than ultrasound. It also provides advanced mechanisms to image the micro-structure and function of the fetal brain \textit{in-utero}~\cite{gholipour2014fetal}. Hence, it provides information that cannot be obtained by any other imaging technique. Fetal MRI has enabled \textit{in-vivo} mapping and quantitative analysis of cortical maturation in the fetus~\cite{clouchoux2012quantitative,rajagopalan2011local,scott2011growth,corbett20113d,wright2014automatic,vasung2019quantitative,vasung2020spatiotemporal}, and has shed light on patterns and forces that govern cortical folding and expansion~\cite{kroenke2018forces,wang2017folding, llinares2019deconstructing, rana2019subplate}. It has also enabled studies that have quantified altered cortical development in fetuses with congenital disorders of the brain and heart~\cite{clouchoux2013delayed,im2017quantitative,tarui2018disorganized,benkarim2018cortical,ortinau2019early}. These studies require accurate segmentation of the developing fetal brain tissue, in particular the cortical plate, on fetal MRI.
	
	Fetal brain tissue segmentation on MRI has been historically challenged due to the limitations of imaging the fetal brain \textit{in-utero} by MRI. In particular, fetal movements disrupt the spatial encoding that is necessary to acquire real 3D images of the fetal brain. Therefore, fetal MRI is performed through 2D stack-of-slice acquisitions that do not make 3D images with coherent anatomic boundaries in 3D. Additionally, there are technical challenges in acquiring 3D fetal brain MRI scans. Hence, there has been a lack of carefully-labeled 3D (and more importantly spatiotemporal or 4D) images of the fetal brain MRI that can be used as atlases for multi-atlas automatic segmentation, or as training data for learning-based segmentation methods. These issues, however, have been addressed in the recent years through the development of super-resolution fetal brain MRI reconstruction methods, e.g.~\cite{gholipour2010robust,studholme2011mapping,kuklisova2012reconstruction,kainz2015fast,alansary2017pvr,ebner2020automated}. These methods, in-turn, led to the construction of carefully labelled, spatiotemporal atlases of fetal brain MRI~\cite{gholipour2017normative}. These developments allowed automatic segmentation of fetal brain tissue, mainly through multi-atlas segmentation methods~\cite{habas2010atlas,gholipour2017normative,makropoulos2018review}. Nonetheless, accurate segmentation of the fetal brain tissue, in particular segmentation of thin structures such as the cortical plate (CP), remains challenging due to several outstanding issues that we discuss next. 
	
	\subsection{Fetal Cortical Plate Segmentation}
	
	The CP was manually or semi-automatically segmented on MRI in early studies on quantitative analysis of fetal cortical maturation~\cite{clouchoux2012quantitative,clouchoux2013delayed,im2017quantitative}. Manual segmentation of the super-resolved reconstructed fetal brain MRI scans, however, is laborious and time consuming~\cite{gholipour2017normative}. Therefore, more recent studies relied upon careful manual refinement of automatic (multi-atlas based) fetal brain MRI segmentations~\cite{ortinau2019early,vasung2019quantitative}. Nonetheless, manual refinement of automatic segmentations is also tedious and difficult on super-resolved 3D images. 
	
\begin{figure}[htbp]
		\centering
		\includegraphics[width=0.44\textwidth]{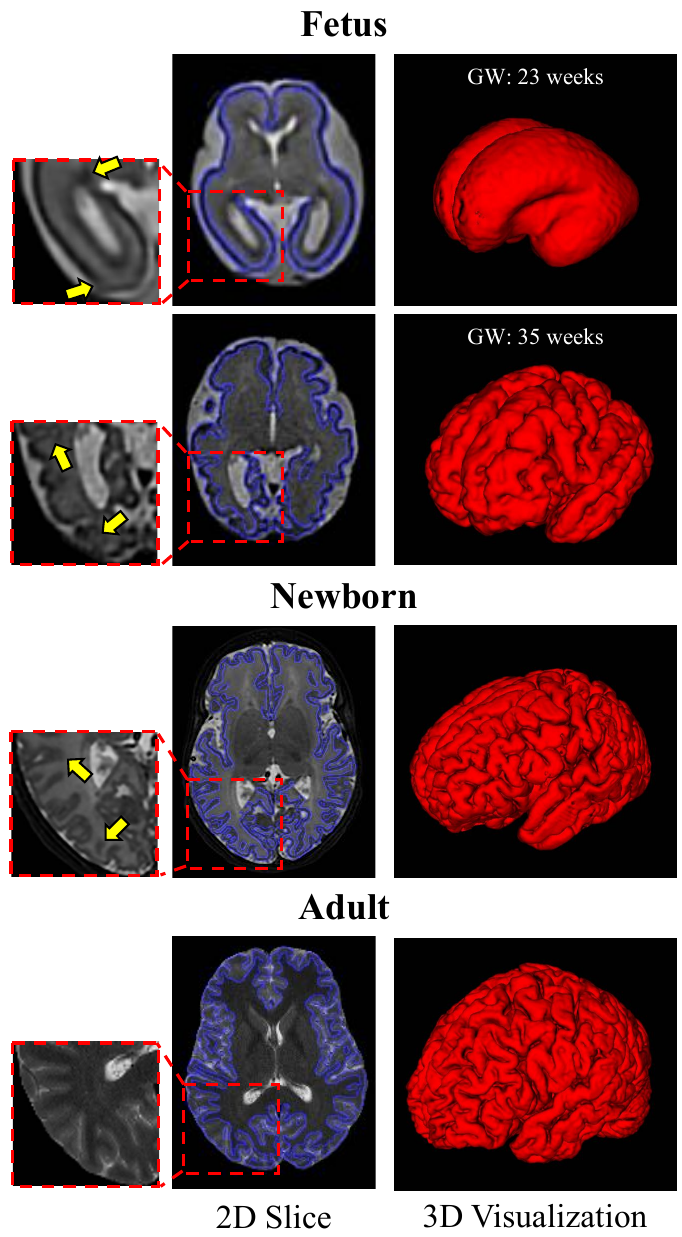}
		\caption{Comparing the appearance of the fetal brain (in the second and third trimesters) with the brain of a newborn and an adult on T2-weighted MRI. The first column shows an axial slice of brain MRI where the blue line shows the boundary of the cortical plate. The second column is the 3D surface mesh visualization of the CP. Compared to the newborn and adult brains, the fetal brain exhibits wide variations in size and morphology. In addition, the relatively low spatial resolution of fetal MRI, the thin size of the CP, and the reversed contrast of the GM and WM that causes WM-like intensities in the CP-CSF interface (due to partial voluming) (shown by arrows), pose significant challenges for automatic CP segmentation. GW: Gestational Weeks.}
		\label{fig:difference_brain}
	\end{figure}

	Cortical plate segmentation on fetal MRI is particularly challenging as the fetal CP is a very thin ribbon with a thickness that is comparable to the best achievable resolution on fetal MRI scans. As a result, partial voluming effects cause major issues in fetal CP segmentation on MRI. Furthermore, because the tissues are immature, the contrast of the developing white matter (WM) and gray matter (GM) for the fetal (and newborn) brain on MRI is the reverse of the contrast of the WM and GM on MRI of mature brains~\cite{xue2007automatic,weisenfeld2009automatic}. This exacerbates the partial voluming effect as partial voluming in the CP and the cerebrospinal fluid (CSF) interface generates a WM-like intensity that confuses automatic segmentation algorithms. This can cause registration errors and inaccurate segmentations that appear as holes and topological errors in thin or highly-folded areas of the fetal cortex. This issue can be seen in Fig.~\ref{fig:difference_brain}, which shows T2-weighted MRI images of the brain of two fetuses scanned at 23 weeks and 35 weeks gestational age as well as T2-weighted brain MRI images of a newborn and an adult. Compared to the adult brain, in which the GM intensity falls between the range of the intensities of the WM and CSF, in the fetal and newborn brains the WM appears brighter than the CP. This is problematic as the partial voluming between CP and CSF resembles WM, which may then result in topological errors in segmentation.
	
	Another factor that makes automatic segmentation of the fetal CP challenging is the substantial variations in fetal brain morphology due to the rapid development of the brain throughout gestation. As Fig.~\ref{fig:difference_brain} shows, the CP of the fetal brain in the second trimester (e.g. at 23 weeks) has a smooth shape and does not have many sulci. With the rapid growth of the fetus, sulci form and the CP folds, ultimately forming an adult-like, highly-folded area just before birth, which then continues to mature after birth. The large variations in morphology, appearance, and scale of the fetal CP, along with the challenges discussed earlier, make it difficult to develop effective and robust solutions for automatic CP segmentation in fetal MRI.
	
	\subsection{Related Works}
	
	For a review of automatic fetal and neonatal brain MRI segmentation techniques we refer to~\cite{makropoulos2018review}. In our review of the related works here we focus on techniques that segment the fetal brain \textit{tissue} on \textit{3D reconstructed} fetal brain MRI images. Therefore, we do not review techniques that only segment (i.e., extract) the entire fetal brain on the original (stack-of-slice) fetal MRI scans. Those techniques that extract and segment the entire fetal brain on slices of the original MRI scans are useful as a pre-processing step for slice-by-slice motion correction and 3D super-resolution reconstruction of fetal brain MRI. Our work, on the other hand, is focused on post-reconstruction analysis of fetal brain cortical development. 
	
	The early works on 3D fetal brain MRI tissue segmentation were based on deformable atlas-to-image registration and propagation of labels from MRI atlases to MRI scans of query subjects. Habas \textit{et al.} \cite{habas2008atlas} first proposed an atlas-based solution to segment the fetal brain tissue with a focus on the germinal matrix. Inspired by~\cite{habas2008atlas}, they incorporated a depth-based geometric representation model into the conventional atlas segmentation strategy~\cite{habas2008atlas} by providing additional anatomical constraints~\cite{habas2009statistical}. In their follow-up work, Habas \textit{et al.} constructed 3D statistical MRI atlases of fetal brain tissue distribution~\cite{habas2010spatiotemporal} and used them to segment fetal brain tissue on 3D reconstructed fetal brain MRI images~\cite{habas2010atlas}.
	
	Serag \textit{et al.}~\cite{serag2012multi} built a spatiotemporal atlas of the fetal brain from reconstructed fetal brain MRI scans of 80 fetuses scanned in the gestational age (GA) range of 23 to 37 weeks.\footnote{\url{http://brain-development.org/brain-atlases/fetal-brain-atlases/}} They used the atlas in a multi-atlas deformable registration and label propagation framework to segment fetal brain tissue on reconstructed fetal brain MRI images. With their atlas and segmentation technique they reported average Dice similarity coefficient (DSC) of 0.84 for automatic CP segmentation.
	
	Gholipour \textit{et al.}~\cite{gholipour2012multi} developed a multi-atlas multi-shape segmentation technique based on a combination of label fusion from multiple age-group atlases and shape priors. They used their atlas to segment lateral ventricles in fetuses at different ages with different levels of ventriculomegaly. Gholipour \textit{et al.}~\cite{gholipour2014construction,gholipour2017normative} then developed and distributed fully labeled spatiotemporal (4D) MRI atlases of normal fetal brain growth at 1mm isotropic resolution in the gestational age range of 19-39 weeks.\footnote{\url{http://crl.med.harvard.edu/research/fetal_brain_atlas/}} They used reconstructed images of 81 normal fetuses to construct the atlas and used it for multi-atlas tissue segmentation where they achieved average DSC of 0.9 and 0.84 for CP segmentation (using a leave-one-out strategy) for fetuses scanned at $\leq34$ weeks and $>34$ weeks, respectively.
	
	Khalili \textit{et al.}~\cite{khalili2019automatic} developed a fully convolutional neural network based on the U-Net architecture to segment fetal brain tissue on 3D reconstructed fetal brain MRI scans of 12 fetuses in the 23-35 weeks GA range. They reported average test DSC of 0.835 for CP segmentation, which compared favorably against the average DSCs of 0.82 and 0.84 reported for the different test sets in~\cite{habas2010atlas} and \cite{serag2012multi}, respectively. Although, as noted in~\cite{khalili2019automatic}, these values should not be compared directly because they were calculated on different test sets.
	
	\subsection{Contributions}
	To reduce the burden of manual refinement of segmentation and topology correction, and to improve quantitative analysis of fetal cortical plate development, we aimed to significantly improve automatic segmentation of the cortical plate in fetal MRI. We aimed to address the outstanding issues that were mentioned above, in particular to deal with the substantial variability in the size, shape, and complexity of the thin fetal brain CP. To this end, in this investigation we developed a new network architecture with novel attention modules using mixed kernel convolutions. The attention modules helped our network extract and learn important multi-scale information from the feature maps in a stage-wise manner. We compared our trained model with several state-of-the-art models including: the 3D U-Net~\cite{cciccek20163d} used in fetal brain MRI segmentation~\cite{khalili2019automatic}, the Plane Aggregated U-Net (PAUNet)~\cite{Hong2019FetalCP}, the Attention U-Net~\cite{oktay2018attention}, and the Squeeze $\&$ Excitation Fully Convolutional Network (SE-FCN)~\cite{roy2018recalibrating}. We compared methods quantitatively based on several performance metrics on held-out test sets. We also compared the results of our method against a multi-atlas segmentation technique based on blind expert evaluations. The results indicate significant improvement in CP segmentation in fetal MRI using our proposed network with attentive learning.
	
	The paper is organized as follows: Section~\ref{sec:Methods} involves the materials and methods including the details of the fetal MRI dataset, our proposed network architecture, the attention refinement module, loss function, and training. Section~\ref{sec:results} describes the experimental results; Section~\ref{sec:discussion} includes a discussion; and Section~\ref{sec:conclusion} includes our concluding remarks.
	
	\section{Materials and Methods}
	\label{sec:Methods}
	
	\subsection{Fetal MRI data}
	\label{sec:data}
	The fetal MRI dataset used in this study consisted of super-resolution reconstructed volumes of 57 fetuses scanned at a gestational age (GA) between 16 and 39 weeks (29.5$\pm$5.5). The scans were performed on 3T Siemens Skyra scanners with body matrix and spine coils. These images were reconstructed from slices of T2-weighted single shot fast spin echo scans of each fetus, repeated 3-4 times in each of the axial, coronal, and sagittal planes with respect to the fetal head. The protocol and the research imaging was approved by the institutional review board committee. Written informed consent was obtained from all pregnant women volunteers who participated in the study. The in-plane resolution of the original scans was 1 mm, the slice thickness was 2-3~mm, and the repetition time and echo time were 1500 and 120 ms, respectively.
	
	Approximate fetal brain masks were automatically extracted on the original scans using the model in~\cite{salehi2018real}. They were used along with  slice-to-volume registration (for inter-slice motion correction) to reconstruct a super-resolved 3D volumes of the fetal brain using the algorithm described in~\cite{kainz2015fast} at an isotropic resolution of 0.8 $mm^3$. Final 3D brain masks were generated on the reconstructed images using the Auto-Net~\cite{salehi2017auto}, and were manually corrected in ITK-SNAP~\cite{yushkevich2019user} as needed. Brain-extracted reconstructed volumes were then registered to a spatiotemporal fetal brain MRI atlas and the brain tissue and cortical plate were segmented using the procedure described in~\cite{gholipour2017normative}. To generate reference (ground truth) segmentation data, this procedure involved manual segmentation and refinement of the cortical plate in several rounds in different planes. This procedure took anywhere between 2 to 8 hours depending on the age of the fetus and the quality of the reconstructions.
	
	\subsection{Model Architecture}
	
\begin{figure*}[!htbp]
		\centering
		\includegraphics[width=0.95\textwidth]{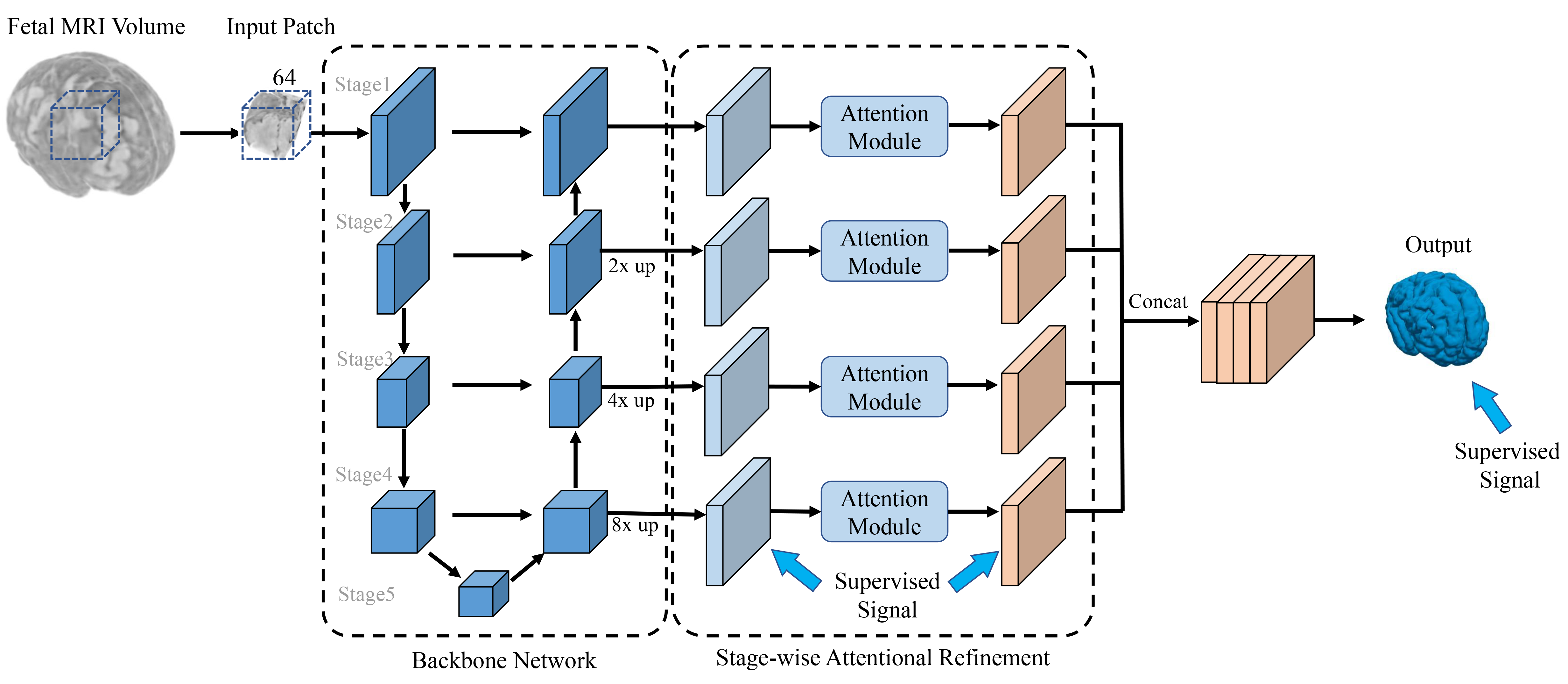}
		\caption{Schematic illustration of our proposed CNN architecture, which consists of a backbone fully convolutional encoder-decoder network with stagewise forward skip connections, and a stagewise attention refinement module. The network takes sliding, overlapping 3D patches of size $64\times64\times64$ from the input image, and using training data, learns to generate a 3D segmentation of the input image.}
		\label{fig:framework}
	\end{figure*}

	Figure~\ref{fig:framework} shows a schematic representation of our proposed deep attentive fully convolutional neural network (CNN) architecture for CP segmentation in fetal MRI. Our model consists of a backbone network with an encoder-decoder architecture with forward skip connections from the encoder stages to the corresponding decoder stages. This is followed by a stage-wise attention refinement module that leverages mixed kernel convolutions to capture multi-scale contextual information. In this section and the following subsections, we explain the details of our model and its attentive learning that enables learning to segment the complex and variable structure of the cortical plate in fetuses scanned at different ages.
	
	As shown in Fig.~\ref{fig:framework}, our network works on 3D patches of size $64^3$. This patch size was chosen in a trade-off to enable learning multi-scale features while limiting memory requirements. For each 3D input patch, the network outputs a CP probability map of the same size in an end-to-end manner. The network first extracts a series of feature maps with different resolutions. The shallower feature maps (Stage 1) contain high-resolution details needed for accurate delineation of the CP boundary. The deeper features maps contain coarse and high-level information that help predict the overall outline of the CP. Our backbone network uses forward skip connections~\cite{cciccek20163d} and convolutions with residual connections~\cite{he2016deep}. The input of each stage is first processed by two $3\times3\times3$ convolutional layers followed by batch normalization (BN)~\cite{ioffe2015batch} and parametric rectified linear unit (PReLU) for activation~\cite{he2015delving}. 
	
	A shortcut (i.e., residual) connection is added between the input and the output of every convolutional block in this network. The numbers of feature maps in the encoder part of the network are 16, 32, 64, 128, 128, increasing the number of features as their size shrinks. We limited the number of feature maps at the last encoder stage to 128 to accommodate training on a graphical processing unit (GPU) with 8GB of memory. Every feature map computed by the backbone network is then upsampled using trilinear interpolation to the size of the input patch. Then a convolutional operation with a kernel size of $1\times1\times1$ is applied to every feature map to create 16 features in each of the feature maps. The feature maps then go through deep supervision modules~\cite{xie2015holistically} that improve the gradient flow and encourage learning more useful representations. Similar deep supervision modules are also used on the outputs of the attention modules and merge the resulting feature maps via concatenation. These feature maps go through two convolutional layers followed by BN and PReLU to produce the CP probability map.

	\subsection{Stage-wise Attentive Refinement}
	\label{attentive_refinement}
	
\begin{figure*}[!htbp]
		\centering
		\includegraphics[width=0.95\textwidth]{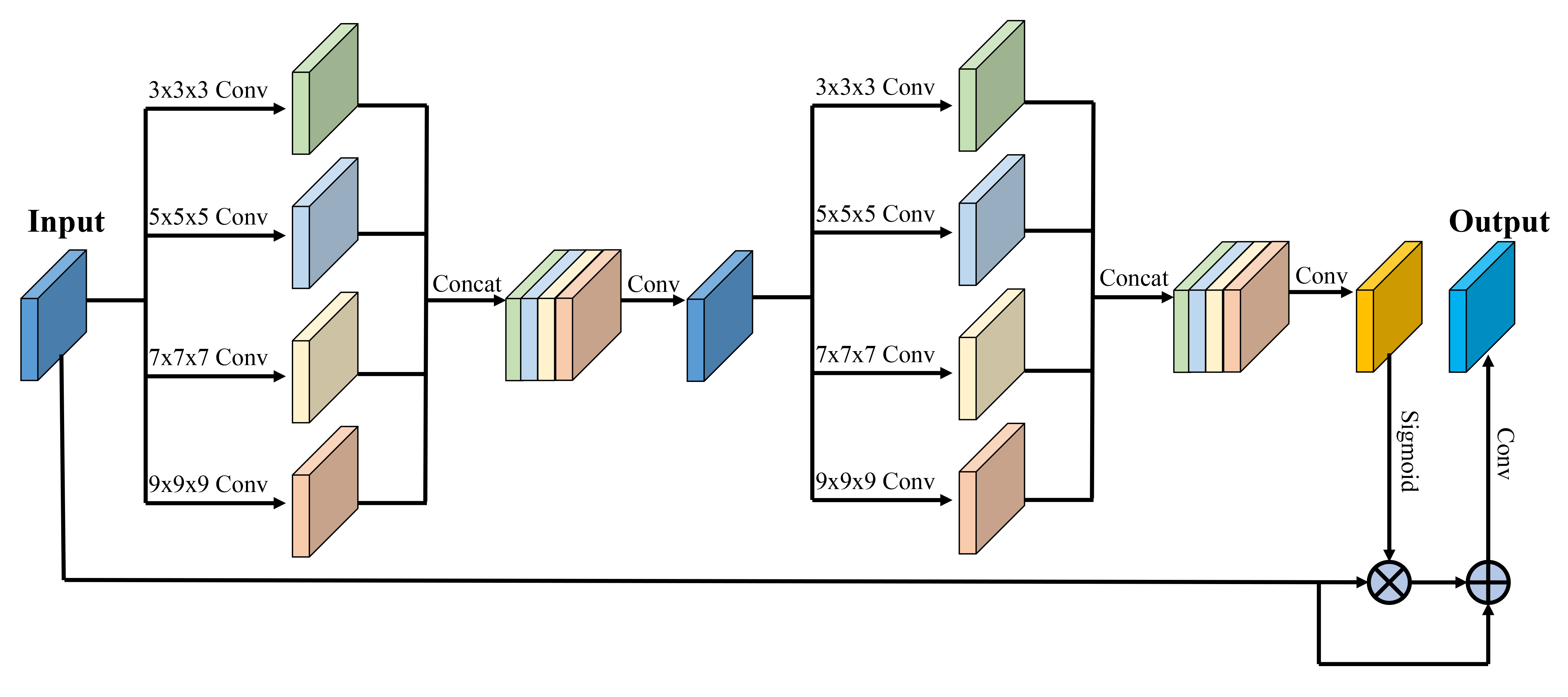}
		\caption{Schematic illustration of the architecture of our attention module, which takes an input feature map and generates a refined attentive feature map of the same size. The module includes two layers of group convolutional blocks followed by feature concatenation and $1\times1\times1$ bottleneck convolution blocks. The module uses a sigmoid activation function and a residual connection between its input and output to reduce the difficulty of learning attentive maps.}
		\label{fig:attention}
	\end{figure*}

	The goal of our attention modules is to increase the network's ability to capture the multi-scale details of the brain CP. We expect that these modules increase the richness of the information of the multi-scale feature maps learned by the backbone network. Figure~\ref{fig:attention} shows the architecture of our proposed attention module as it is used in each of the stages of our stage-wise attentional refinement network in Fig.~\ref{fig:framework}. We explain the details of the attention module here.
	
	As shown in Fig.~\ref{fig:attention}, our attention module takes the feature maps generated from each stage of the backbone network and outputs a refined attentive feature map of the same size. The operation of the attention module at stage $i$ has the following functional form:
	\begin{equation}\label{eq_attention}
		F^{'}_{i} = f_{i}(F_{i};\theta) \otimes F_{i} + F_{i}.
	\end{equation}
	
	\noindent where $\theta$ denotes the parameters of the module that include the weights of the convolutional layers; $F_{i}$ and $F^{'}_{i}$ respectively denote the input feature maps and the output refined attentive feature maps at stage $i$; and $\otimes$ is element-wise multiplication.
	
	In our attention module, the input feature map $F_{i}$ is first processed with group convolutional blocks. The sizes of the kernels vary from $3\times3\times3$ to $9\times9\times9$, but the number of feature maps in each group is fixed at 4. Each convolutional operation is followed by BN and PReLU. This design not only reduces the GPU memory footprint and computation, but also increases the richness of the learned representations as demonstrated by similar designs~\cite{xie2017aggregated}. The generated feature maps are concatenated and passed through a second, and identical, group of convolutional blocks. A convolution layer with a kernel size of $1\times1\times1$ is used to merge these multi-scale feature maps into a single feature map. Finally, a sigmoid activation is applied to obtain the attentive map $A_{i}$. The attentive map is multiplied element-wise with the input feature map to encourage attention to the relevant locations in the feature map. The refined attentive feature map is added to the input feature map in the spirit of residual connections to reduce the difficulty of learning the attentive map. Note that we use BN on the feature maps from these two branches before adding them together. This is necessary to ensure that the values of the two feature maps are not very differently distributed.
	
	\subsection{Loss Function}
	
	We trained our network with the weighted binary cross-entropy loss, $L_{wbce}$:
	\begin{equation}
		L_{wbce} = \frac{1}{N}\sum_{i=1}^{N}[\alpha g_{i}\log{p_{i}} + (1-g_{i})\log{(1-p_{i})}].
	\end{equation}
	
	\noindent In this equation, $N$ is the number of voxels in the patch. Moreover, $p_{i}$ and $g_{i}$ denote the predicted cortical plate probability map and the binary ground truth cortical plate map at voxel $i$. Lastly, $\alpha$ is the weight hyperparameter, which is set independently for each training mini-batch.
	
	Our overall loss function is the weighted sum of the losses at different points in the network involving supervision:
	\begin{equation}
		L_{total} = \sum_{i=1}^{n}w^{i}L_{signal}^{i} + \sum_{j=1}^{n}w^{j}L_{signal}^{j} + w^{p}L_{signal}^{p}.
	\end{equation}
	
	\noindent the above equation, $w^{i}$ and $L_{signal}^{i}$ denote the weight and loss for the points of supervision at stage $i$ of the backbone network, as shown in Fig.~\ref{fig:framework}. Similarly, $w^{j}$ and $L_{signal}^{j}$ denote the weight and loss for the point of supervision at stage $j$ of the attentive refinement network. The number of stages is denoted with $n$, and $w^{p}$ and $L_{signal}^{p}$ denote, respectively, the weight and loss computed at the final network output. Using cross-validation in this study, we set $w^{i}= 0.8, 0.7, 0.6, 0.5|_{i=1:4}$, $w^{j}= 0.8, 0.7, 0.6, 0.5|_{j=1:4}$, and $w^{p}=1$.
	
	\subsection{Implementation and Training}
	
	We implemented our network in PyTorch \cite{paszke2019pytorch} and trained it in an end-to-end manner using Adam optimizer \cite{kingma2014adam} with a weight decay of 1e-4. We used an initial learning rate of 1e-3 and reduced it by a factor of 0.1 after 3k and 4.5k training iterations. Learning stopped after 6k iterations. We used random 3D flip and 90 degree rotations for augmentation of input patches during training. We also implemented and trained alternative networks for comparison with the same training data and with similar hyper-parameters for optimization and data augmentation. We trained all models on an Nvidia GTX-1080Ti GPU with a mini-batch size of 4. Training our network with 6k iterations took approximately 24 hours.
	
	\section{Experiments and Results}
	\label{sec:results}
	
	In this section, we first describe the alternative, state-of-the-art deep learning models that we have compared and the evaluation criteria that we used to compare the results. Then we describe the results of the comparisons as well as expert evaluations in comparing the results of our deep learning technique with multi-atlas segmentation. We discuss the results of the comparisons in detail in Section~\ref{sec:discussion}.
	
	\subsection{Alternative Techniques}
	
	To demonstrate the advantages of our proposed method in fetal brain CP segmentation, we compared it with four alternative deep learning models, including two non-attentive models and two attentive models.
	
	\emph{Non-attentive models:}
	\begin{itemize}
		\item 3D UNet~\cite{cciccek20163d}: is based on the well-known U-Net architecture~\cite{ronneberger2015u}. It has proven to be a powerful segmentation CNN that has achieved success in multiple segmentation tasks, e.g.~\cite{milletari2016v,salehi2017tversky,salehi2018real,du2020medical,minaee2020image}. The 3D UNet leverages skip connections to strengthen the fusion of multi-level features to generate refined segmentation results.
		\item Plane Aggregated U-Net (PAUNet)~\cite{Hong2019FetalCP}: segments the CP with a 2D UNet with dense blocks~\cite{jegou2017one} from each plane and constructs a final prediction via plane aggregation. We used the mean of predictions for plane aggregation. 
	\end{itemize}
	
	\emph{Attentive models:}
	\begin{itemize}
		\item Attention UNet~\cite{oktay2018attention}: integrates a self-attention gating module into the U-Net model, which allows attention coefficients to be more specific to local regions.
		\item Squeeze $\&$ Excitation Fully Convolutional Network (SE-FCN) \cite{roy2018recalibrating}: uses spatial and channel-wise squeeze and excitation attention to improve segmentation performance.
	\end{itemize}
	
    In the ablation studies, we investigated the impact of our proposed attentive module and deep supervision mechanisms. First, to demonstrate the advantages of our stage-wise attentive refinement module, we compared our framework with a model in which we removed the attention module and concatenated all feature maps from the backbone network with different resolutions to form the final prediction. We refer to this model as 3D Deeply Supervised Residual Network (DSRNet). Then, to investigate the efficacy of deep supervision, we trained our model with different supervision strategies. Specifically, we compared training with supervision at: 1) only the final network output, 2) both the final output and the multi-scale features, and 3) the attentive maps instead of attentive features in an approach similar to~\cite{wang2019automatic}.
	
	We also compared the results of our deep learning model with a state-of-the-art multi-atlas segmentation (MAS) technique as described in~\cite{gholipour2017normative}. This technique uses diffeomorphic deformable registration~\cite{avants2011reproducible} between each atlas and the query image to propagate labels from the atlas to the query image and then fuses the label maps using probabilistic STAPLE~\cite{akhondi2013simultaneous}.
	
	\subsection{Evaluation Criteria}
	
	To compare segmentation results quantitatively on the test set, we used three evaluation metrics: 1) the Dice similarity coefficient (Dice), 2) the 95 percent Hausdorff Distance (95HD), and 3) the Average symmetric Surface Distance (ASD)\cite{chang2009performance,mendrik2015mrbrains,yeghiazaryan2015overview}. The Dice is used to estimate the spatial overlap between the prediction and ground truth, 95HD is an outlier-robust measure based on the Hausdorff distance between two boundaries, and ASD is the average of all distances from points in the surface of the prediction to the ground truth.
	
	MAS was used to obtain the very initial segmentations for our fetal MRI datasets that experts used as guidance in segmentations~\cite{gholipour2017normative}. Even after significant time-consuming manual segmentations and refinements, we were unable to ensure that the quantitative comparison of our method with MAS based on the overlap and distance metrics was unbiased. Therefore, for comparing our method with MAS, we performed blind expert evaluations. For this purpose, on a set of 15 fetal MRI scans from the test set, three experts independently compared segmentation results of our method and MAS. This was done in a randomized and blind fashion. Specifically, the experts visually compared the two segmentation results overlaid on the grayscale anatomical image of each fetus in different planes. They indicated if either of the segmentations A or B was better or if there were about the same quality.
	
	\subsection{Evaluation Results}
	
	As summarized in Table \ref{tab:results}, our method outperformed other methods based on almost all evaluation metrics. Our method achieved mean Dice of 0.87, 95HD of 0.96, and ASD of 0.28. The best value of each performance measure has been highlighted by boldface text in the table. 
	According to these results our method generated better segmentations than both 3D UNet and DSRNet according to both Dice and ASD, and it outperformed PAUNet based on Dice. Our method showed an average of 5$\%$ and 0.12 mm improvement in Dice and 95HD over PAUNet, respectively. 
	Moreover, our method achieved superior segmentation accuracy compared with the two competing attentive models (Attention U-Net and SE-FCN) in terms of Dice, 95HD, and ASD.
	Comparing the results of the DSRNet with those of the 3D UNet shows the improvement achieved by our backbone network (DSRNet) that exploited residual learning with deep supervision. Comparing our method with DSRNet shows the contribution of our attentive refinement to the improvement in performance. 
    
    In Table~\ref{tab:results}, we have also reported the number of parameters and running time for all compared models. We observe that by adding 2.6$\%$ extra parameters, our attention module could improve the performance by 3$\%$ (from 0.84 by DSRNet to 0.87 by Ours) in terms of Dice. Although our method takes longer than the other methods in inference, all the test times are reasonable for the segmentation of post-acquisition reconstructed images, as these run times are an order of magnitude smaller than the time it takes to reconstruct images.
    	
	\begin{table*}[h] 
		\centering
		\caption{Comparing segmentation results of different methods (3D UNet, PAUNet, Attention UNet, SE-FCN, DSRNet, and our method) based on quantitative performance measures (Dice, 95HD, and ASD) and computational cost (Num of Params, and Run time). The best value for each evaluation measure has been shown in boldface text.}
		\label{tab:results}
		\begin{tabular}{c|c|c|c|c|c} 
			\hline
			\hline
			Method & Dice$\uparrow$ & 95HD(mm)$\downarrow$ & ASD(mm)$\downarrow$ & Num of Params & Run time(s)\\ 
			\hline
			3D UNet & 0.81$\pm$0.06 &1.30$\pm$0.55 &0.54$\pm$0.18 &2.51M &5.70\\
			\hline
			PAUNet\cite{Hong2019FetalCP}\ &0.82$\pm$0.09 &1.08$\pm$0.51 &\textbf{0.28$\pm$0.13} &0.76M &5.03\\
			\hline
			Attention U-Net~\cite{oktay2018attention} &0.85$\pm$0.08 &1.00$\pm$0.58 &0.45$\pm$0.32 &3.01M &8.63\\
			\hline
			SE-FCN~\cite{roy2018recalibrating} &0.82$\pm$0.07 &1.31$\pm$0.82 &0.45$\pm$0.26 &2.59M &8.70\\
			\hline
			DSRNet &0.84$\pm$0.06 &1.12$\pm$0.48 &0.40$\pm$0.17 &2.68M &12.89\\
			\hline
			Ours &\textbf{0.87$\pm$0.06} &\textbf{0.96$\pm$0.38} &\textbf{0.28$\pm$0.14} &2.75M & 45.14\\
			\hline
			\hline
		\end{tabular}
	\end{table*}
	
	To investigate if the differences between methods were statistically significant, we conducted paired \textit{t}-tests between the results of our method and those of 3D UNet, PAUNet, Attention UNet, SE-FCN and DSRNet. This test was carried out for all performance measures (Dice, 95HD, and ASD). In our paired \textit{t}-tests, the significance level was set as 0.05. The \textit{p}-values for the paired t-tests is summarized in Table~\ref{tab:signifcance}. Overall, the results of the comparisons and tests in Tables \ref{tab:results} and \ref{tab:signifcance} indicate that our method performed best among all the techniques that we implemented and examined.
	
	Table~\ref{tab:supervision_loss} shows the results of our experiments to investigate the effectiveness of our deep supervision strategy. In these experiments, we compared our deep supervision strategy with alternative training strategies that apply supervision at different network layers, as explained above. We can observe from Table~\ref{tab:supervision_loss} that our deep supervision strategy (backbone + attentive features + output) achieved the highest segmentation accuracy. The results suggest that supervision at the attentive features level (SAF) achieves slightly better segmentation accuracy than supervision at the attention map level (SAM). Feature maps at different channels contain different representations and require different attention maps. SAM imposes the same attention coefficients on feature maps of different channels, which is not as accurate as SAF for different channels.
	
	\begin{table}[h] 
		\centering
		\caption{\textit{p}-values of paired \textit{t}-tests between the results of each method and our method for the three segmentation performance measures used in this study (Dice, 95HD, and ASD).} 
		\label{tab:signifcance}
		\begin{tabular}{c|c|c|c} 
			\hline
			\hline
			Metrics & Dice & $95$HD & ASD \\ 
			\hline
			3D U-Net vs. Ours & $10^{-5}$ &$10^{-3}$ &$10^{-10}$\\
			\hline
			PAUNet vs. Ours &$10^{-3}$ &0.21 &0.92 \\
			\hline
			Attention UNet vs. Ours &0.12 &0.70 &$10^{-3}$\\
			\hline
			SE-FCN vs. Ours &$10^{-3}$ &0.01 &$10^{-4}$\\
			\hline
			DSRNet vs. Ours &0.02 &0.09 &$10^{-3}$\\
			\hline
			\hline
		\end{tabular}
	\end{table}
	
	\begin{table*}[h]
		\centering
		\caption{Supervised learning based on different features.}
		\label{tab:supervision_loss}
		\begin{tabular}{c|c|c|c|c|c|c}
			\hline
			\hline
			\multicolumn{4}{c|}{Supervision} & \multirow{2}{*}{} & \multirow{2}{*}{} & \multirow{2}{*}{} \\ \cline{1-4}
			Backbone & Attentive Feature & Attention Map & Output & Dice$\uparrow$ & 95HD(mm)$\downarrow$ & ASD(mm)$\downarrow$  \\ \hline
			&  &   & \checkmark & 0.85$\pm$0.06  & 1.14$\pm$0.57  & 0.49$\pm$0.18 \\ \hline
			\checkmark &  &  & \checkmark & 0.86$\pm$0.06  & 1.12$\pm$0.43 &0.46$\pm$0.38 \\ \hline  
			\checkmark &  &\checkmark  & \checkmark & 0.86$\pm$0.05  & 1.07$\pm$0.34  & 0.32$\pm$0.12 \\ \hline   
			\checkmark &\checkmark  &  & \checkmark & \textbf{0.87$\pm$0.06} & \textbf{0.96$\pm$0.38}  & \textbf{0.28$\pm$0.14} \\ \hline   
			\hline
		\end{tabular}
	\end{table*}
	
	\begin{figure*}[!htbp]
		\centering
		\includegraphics[width=0.8\textwidth]{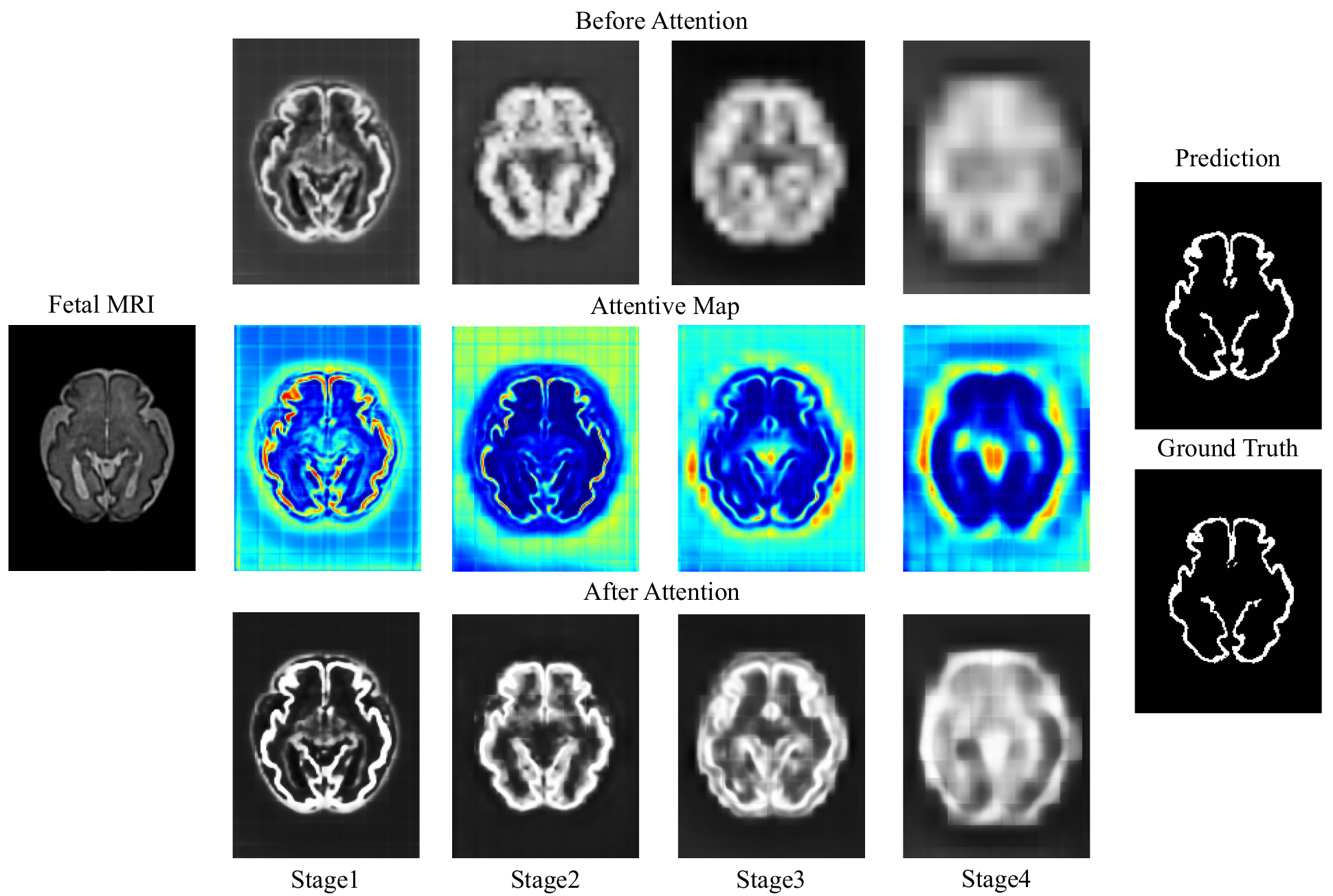}
		\caption{The illustration of the feature maps in the attention module. The first column shows an original 3D reconstructed fetal brain MRI image. The second to the fifth columns show the feature maps in the attention module from stage1 to stage4. The first row to the third row shows the input features, the attentive maps, and the output features of the attention module, respectively. The third column shows the prediction of the network and the ground truth.}
		\label{fig:attention_map}
	\end{figure*}
	
\begin{figure*}[!htbp]
		\centering
		\includegraphics[width=0.95\textwidth]{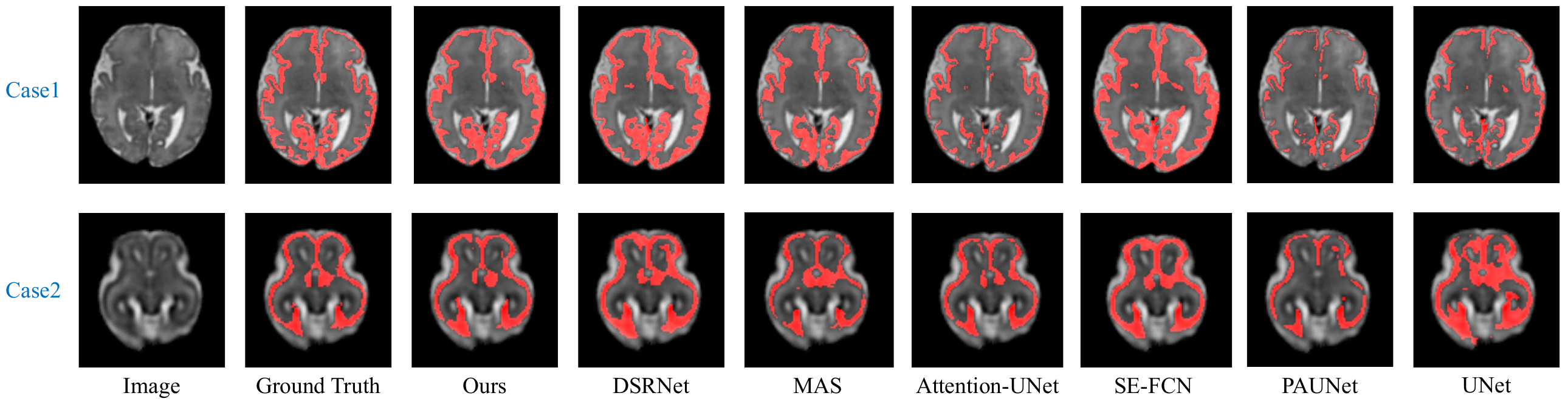}
		\caption{Visual comparison of the segmentation results obtained from different methods (ours, DSRNet, MAS, PAUNet, and 3D UNet) compared to the ground truth for two representative test cases. As supported by the results in Table~\ref{tab:results}, our method generated segmentations that were most similar to the ground truth.}
		\label{fig:2d_segmentation}
	\end{figure*}

	Figure~\ref{fig:attention_map} illustrates the feature maps before and after attentive refinement and the corresponding attentive map. The attentive maps highlight the regions that are important for the task (i.e cortical plate). Figure~\ref{fig:2d_segmentation} shows axial slices of two representative cases with CP segmentation overlaid on the grayscale image using different methods, compared to the ground truth. In terms of the topology of CP segmentations, these results show that 1) the PAUNet and the UNet under- and over-segmented different regions of the CP depending on contrast and partial voluming; 2) the Attention U-Net under-segmented the CP; 3) the SE-FCN over-segmented the CP; and 4) our attentive model generated segmentations that were better than our DSRNet and were the most similar (among all models) to the ground truth. Visual comparison of the DSRNet results and the results of our method shows the extents of the differences in CP segmentation associated with the differences in quantitative measures reported in Tables \ref{tab:results} and \ref{tab:signifcance}.
	
\begin{figure*}[!htbp]
		\centering
		\includegraphics[width=0.95\textwidth]{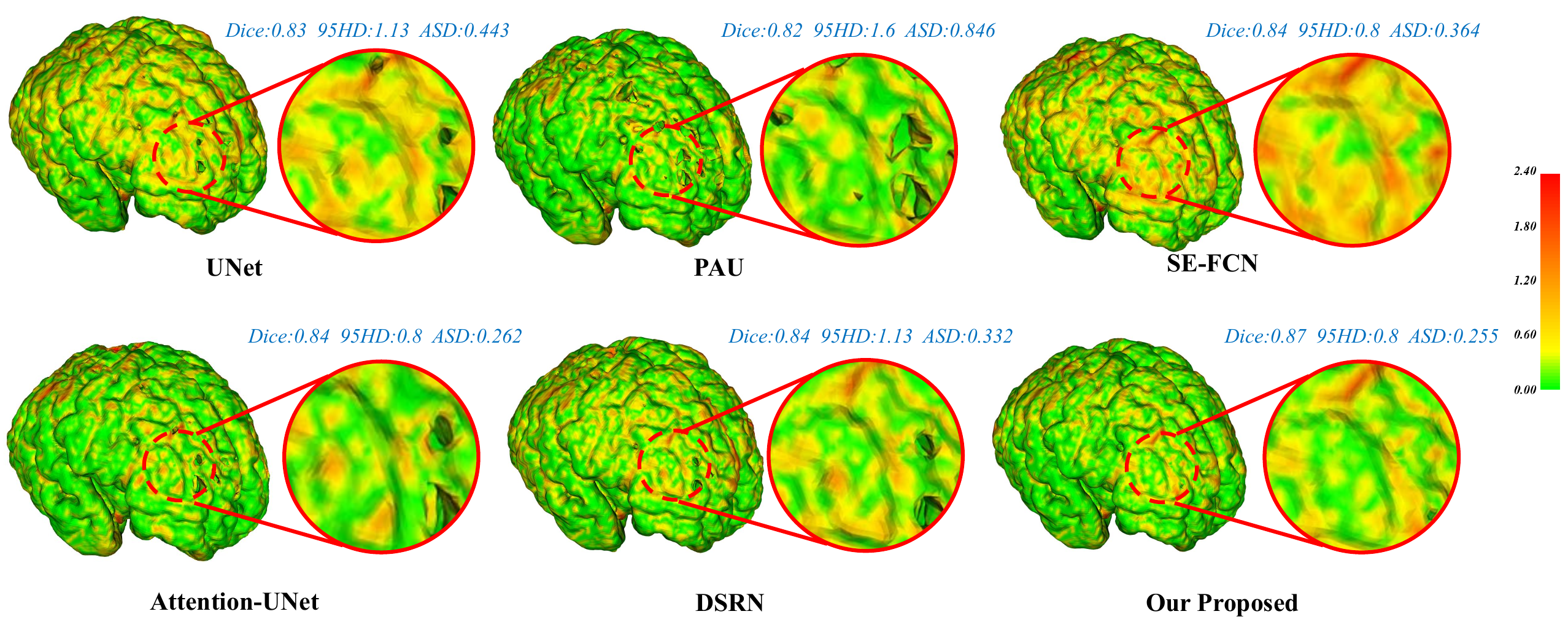}
		\caption{Visualization of the cortical surface error maps on a representative test case along with the values of the performance measures for different methods (3D UNet, PAUnet, SE-FCN, Attention U-Net, DSRNet, and our method). Our method generated the lowest average error, whereas other methods generated holes in the CP surface.}
		\label{fig:surface_error}
	\end{figure*}

	Figure~\ref{fig:surface_error} shows 3D surface error maps of the CP segmentation generated by each method with reference to the ground truth for a representative test subject. The average values of the performance metrics (Dice, 95HD, and ASD) for this case have also been shown for each method. This shows how the extents of the differences in these values correspond to 3D surface map and topological errors in CP segmentation. We can observe that our method achieved the most accurate segmentation results among all the compared methods. Our method generated topologically more accurate CP segmentations than the other methods. In particular, there exist many holes in the segmentation results of UNet, PAUNet, DSRNet, and Attention U-Net in areas with significant partial voluming. This occurred due to the inability of those techniques in leveraging the useful information of multi-scale features. The PAUNet generated low surface errors in many areas, but completely failed in other areas. The SE-FCN generated high surface errors in many areas. Our DSRNet, overall, performed much better than the other competing methods and its performance significantly improved through deep attentive learning as can be seen in the results of our method in the lower right side of Fig.~\ref{fig:surface_error}. 
	
	\subsection{Expert Evaluation Results}
	Table~\ref{tab:voting results} shows the expert evaluation results of CP segmentation using our method against MAS (multi-atlas segmentation) for 15 test subjects. We separated the test subjects to 9 and 6 based on the original reconstructed image quality. This was done because we expected the segmentation to be more challenging in low-quality reconstructions due to the effects of residual motion artifacts and partial voluming. The last row of the table shows the results of majority voting among the three experts, where a three-way tie went to equal. This evaluation shows that our method was chosen most often as the more accurate segmentation method by each expert and also based on majority voting. Our method outperformed MAS regardless of the quality of the input reconstructed image, which shows its robust and accurate performance. It should also be noted that our method generated the CP segmentation for every test subject in less than 1 minute. MAS, on the other hand, took approximately 20 minutes because of the need for deformable registration between each atlas and the subject. 
	
	\begin{table}[h] 
		\centering
		\caption{Expert evaluation results of comparing CP segmentations using our method against multi-atlas segmentation (MAS).} 
		\label{tab:voting results}
		\begin{tabular}{c|c|c|c|c|c|c} 
			\hline
			\hline
			\multirow{2}{*}{Voter}&
			\multicolumn{3}{c|}{Good Reconstruction}&\multicolumn{3}{|c}{Bad Reconstruction}\cr\cline{2-7}
			
			& Ours & MAS & Equal & Ours & MAS & Equal\cr 
			\hline
			Expert1 &9/9 &0/9 &0/9  &6/6 &0/6 &0/6\cr
			\hline
			Expert2 &5/9 &1/9 &3/9  &3/6 &1/6 &2/6\cr
			\hline
			Expert3 &9/9 &0/9 &0/9 &4/6 &2/6 &0/6\cr
			\hline
			Majority &9/9 &0/9 &0/9 &4/6 &0/6 &2/6\cr
			\hline
			\hline
		\end{tabular}
	\end{table}

	\section{Discussion}
	\label{sec:discussion}
	In this paper we presented a deep attentive convolutional neural network for cortical plate segmentation on fetal MRI. Fetal cortical plate segmentation in MRI is very challenging due to the rapid changes and variations in the microstructure and shape of the transient fetal brain compartments. The relatively low spatial resolution of fetal MRI compared to the thin structure of the CP makes the task more challenging. Due to the large variations in CP structure and severe partial voluming effects, automatic segmentation methods struggle to extract multi-scale contextual information from images. To alleviate those issues, our proposed network is equipped with stage-wise attentive refinement modules with mixed kernel convolutions to capture multi-scale information. Besides, the patch-wise input provides sufficient training data in comparison to the image-wise input, which alleviates the risk of over-fitting~\cite{hesamian2019deep}. 
	
	The 3D UNet and PAUNet are powerful models. However, they do not have specific multi-scale architecture design and only use single scale kernels to generate feature maps. As shown in Figure \ref{fig:2d_segmentation} and \ref{fig:surface_error}, the subtle and highly variable boundaries of the fetal CP were much more effectively learned and predicted through the multi-scale attentive modules that exploited different kernel widths in our network. Compared with the 3D UNet and PAUNet, our ablation study method (DSRNet) achieved better segmentation of the cortical regions by concatenating multi-scale feature maps. However DSRNet still lost some details in the subtle boundary of the fetal CP. Our method exploited group convolutional blocks with multiple kernel sizes to extract multi-scale information and generate attentive maps that refined the feature maps at different scales. Using the residual spatial attention mechanism, our attentive network efficiently and effectively aggregated complementary multi-scale information for accurate representation of the fetal CP regions. Given the multi-scale information selected by the stage-wise attentive refinement module, features of highly-variable anatomy were captured accurately and represented effectively during the learning process.
	
	Prior studies have proposed a number of different attention mechanisms for such applications as image classification \cite{hu2018squeeze} and image segmentation \cite{chen2016attention}. Those attention mechanisms have different forms, such as spatial attention \cite{woo2018cbam}, channel-wise attention \cite{hu2018squeeze}, and self-attention \cite{zhang2018self}. However, none of the existing attention mechanisms was suitable for our application. Specifically, channel-wise attention lacks the spatial attention, and self-attention is not suitable for 3D images due to its excessive memory requirements. Our comparisons also showed that our proposed model was superior to prior attentive models for medical image segmentation \cite{oktay2018attention, roy2018recalibrating}. 
	We attribute this to the superior ability of our model to learn multi-scale representations of the cortical plate compared to those other models. Our spatial attention mechanism effectively refined the feature maps from input 3D image patches at different scales to improve segmentation performance.
	
	The idea of using convolutional operations with multiple kernel sizes has been promoted by various network architectures, including Inception \cite{szegedy2015going}, MixNet \cite{tan2019mixconv}, and PSPNet \cite{zhao2017pyramid}. All of these architectures extract multi-scale feature maps using multi-branch convolutional operations with different kernel sizes. These convolutional operations are interlaced with pooling operations to increase the richness of the learned representations. Our attention module is based on similar principles. It consists of multiple convolutional blocks with different kernel sizes to generate multi-scale attention maps.
	
	We used 3D convolutional kernels to build our 3D CNN. 3D CNNs are demanding in terms of GPU memory usage. A common practice to reduce memory requirements in 3D CNNs is to use large 3D patches as inputs. However, if the patches are too small compared to the image structures, models may not learn global context information. In our setting, we were able to use a large input patch size of $64 \times 64 \times 64$ to fit our model on a GPU with 11GB of memory. This patch size covered about $50 \times 50 \times 50~mm^3$ physical volume size, which was sufficiently large compared to the size of the fetal brain.
	
	Atlas-based segmentation techniques are also challenged by the same issues that make learning-based segmentation of the fetal CP challenging. By using age-matched atlases to segment a test subject, these techniques may handle the variability in cortical folding and maturation levels relatively well. However, they are more sensitive to artifacts and partial voluming effects. These techniques rely on deformable registration between each atlas and the test image to propagate the CP label from the atlas to the test subject image. The registration may fail or may be inaccurate in thin and folded areas of the CP where partial voluming disrupts the normal range of the intensity values and features that are used to inform deformable registration. In MAS, registration errors translate into segmentation errors. Another issue with MAS methods is their computational cost and time, which is high due to the required optimization-based deformable registration between multiple atlases and the subject at test time. Once our model is trained, it generates CP segmentation of a test case in less than 1 minute which is an order of magnitude faster than MAS. More importantly, our method generated more accurate segmentations than MAS according to expert evaluations (Table~\ref{tab:voting results}). We attribute this to the superior ability of our model in capturing multi-scale information and learning the context from training samples.
	
	Cortical plate segmentation on reconstructed fetal MRI is a fundamental first step in quantitative analysis of fetal cortical maturation. To evaluate the generalization ability of our model with respect to the fetal MRI image reconstruction algorithms, we reconstructed five fetal MRI scans with NiftyMIC~\cite{ebner2020automated}, and tested them with our model which was trained with images reconstructed using the algorithm by Kainz et al.~\cite{kainz2015fast}. The results have been shown in the Supplementary Figure S1 and Table S1. For this comparison, because our ground truth segmentations were made on images reconstructed using \cite{kainz2015fast}, we had to register and resample the NiftyMIC reconstructions to the reconstructions by \cite{kainz2015fast} for each subject. We note that resampling degrades image quality. Moreover, any registration errors would lead to reduced Dice and increased ASD computations due to misalignment between the ground truth and the test CP segmentations. Despite these factors, the results of this experiment show that our model performed well in the new (NiftyMIC reconstruction) domain.
	
	\section{Conclusion}
	\label{sec:conclusion}
	In this study, we developed a deep attentive neural network with mixed kernel convolutions for automatic cortical plate segmentation in fetal MRI. The key feature of our technique is to leverage a residual spatial attention mechanism to capture multi-scale information. We used two group convolutional blocks with mixed kernels to generate attentive maps to adaptively select and learn important information through the attention mechanism in a deep learning model that leveraged multi-stage deep supervision in both feature extraction and attention modules. Through this novel architecture and the attention mechanism we developed the first 3D fully convolutional deep neural network model for automatic cortical plate segmentation in fetal MRI. Our model outperformed several state-of-the-art deep learning models as well as a multi-atlas segmentation method.

	\bibliographystyle{IEEEtran}
	\bibliography{references-copy.bib}

\begin{thebibliography}{10}
\providecommand{\url}[1]{#1}
\csname url@samestyle\endcsname
\providecommand{\newblock}{\relax}
\providecommand{\bibinfo}[2]{#2}
\providecommand{\BIBentrySTDinterwordspacing}{\spaceskip=0pt\relax}
\providecommand{\BIBentryALTinterwordstretchfactor}{4}
\providecommand{\BIBentryALTinterwordspacing}{\spaceskip=\fontdimen2\font plus
\BIBentryALTinterwordstretchfactor\fontdimen3\font minus
  \fontdimen4\font\relax}
\providecommand{\BIBforeignlanguage}[2]{{%
\expandafter\ifx\csname l@#1\endcsname\relax
\typeout{** WARNING: IEEEtran.bst: No hyphenation pattern has been}%
\typeout{** loaded for the language `#1'. Using the pattern for}%
\typeout{** the default language instead.}%
\else
\language=\csname l@#1\endcsname
\fi
#2}}
\providecommand{\BIBdecl}{\relax}
\BIBdecl

\bibitem{coakley2004fetal}
F.~V. Coakley, O.~A. Glenn, A.~Qayyum, A.~J. Barkovich, R.~Goldstein, and R.~A.
  Filly, ``Fetal {MRI}: a developing technique for the developing patient,''
  \emph{American Journal of Roentgenology}, vol. 182, no.~1, pp. 243--252,
  2004.

\bibitem{gholipour2014fetal}
A.~Gholipour, J.~A. Estroff, C.~E. Barnewolt, R.~L. Robertson, P.~E. Grant,
  B.~Gagoski \emph{et~al.}, ``Fetal {MRI}: a technical update with educational
  aspirations,'' \emph{Concepts in Magnetic Resonance Part A}, vol.~43, no.~6,
  pp. 237--266, 2014.

\bibitem{clouchoux2012quantitative}
C.~Clouchoux, D.~Kudelski, A.~Gholipour, S.~K. Warfield, S.~Viseur,
  M.~Bouyssi-Kobar \emph{et~al.}, ``Quantitative in vivo {MRI} measurement of
  cortical development in the fetus,'' \emph{Brain Structure and Function},
  vol. 217, no.~1, pp. 127--139, 2012.

\bibitem{rajagopalan2011local}
V.~Rajagopalan, J.~Scott, P.~A. Habas, K.~Kim, J.~Corbett-Detig, F.~Rousseau
  \emph{et~al.}, ``Local tissue growth patterns underlying normal fetal human
  brain gyrification quantified in utero,'' \emph{Journal of neuroscience},
  vol.~31, no.~8, pp. 2878--2887, 2011.

\bibitem{scott2011growth}
J.~A. Scott, P.~A. Habas, K.~Kim, V.~Rajagopalan, K.~S. Hamzelou, J.~M.
  Corbett-Detig \emph{et~al.}, ``Growth trajectories of the human fetal brain
  tissues estimated from 3d reconstructed in utero {MRI},'' \emph{International
  Journal of Developmental Neuroscience}, vol.~29, no.~5, pp. 529--536, 2011.

\bibitem{corbett20113d}
J.~Corbett-Detig, P.~Habas, J.~A. Scott, K.~Kim, V.~Rajagopalan, P.~McQuillen
  \emph{et~al.}, ``3d global and regional patterns of human fetal subplate
  growth determined in utero,'' \emph{Brain Structure and Function}, vol. 215,
  no. 3-4, pp. 255--263, 2011.

\bibitem{wright2014automatic}
R.~Wright, V.~Kyriakopoulou, C.~Ledig, M.~A. Rutherford, J.~V. Hajnal,
  D.~Rueckert \emph{et~al.}, ``Automatic quantification of normal cortical
  folding patterns from fetal brain {MRI},'' \emph{NeuroImage}, vol.~91, pp.
  21--32, 2014.

\bibitem{vasung2019quantitative}
L.~Vasung, C.~K. Rollins, H.~J. Yun, C.~Velasco-Annis, J.~Zhang, K.~Wagstyl
  \emph{et~al.}, ``Quantitative in vivo {MRI} assessment of structural
  asymmetries and sexual dimorphism of transient fetal compartments in the
  human brain,'' \emph{Cerebral Cortex}, vol.~30, no.~3, pp. 1752--1767, 2020.

\bibitem{vasung2020spatiotemporal}
L.~Vasung, C.~K. Rollins, C.~Velasco-Annis, H.~J. Yun, J.~Zhang, S.~K. Warfield
  \emph{et~al.}, ``Spatiotemporal differences in the regional cortical plate
  and subplate volume growth during fetal development,'' \emph{Cerebral
  Cortex}, 2020.

\bibitem{kroenke2018forces}
C.~D. Kroenke and P.~V. Bayly, ``How forces fold the cerebral cortex,''
  \emph{Journal of Neuroscience}, vol.~38, no.~4, pp. 767--775, 2018.

\bibitem{wang2017folding}
X.~Wang, C.~Studholme, P.~L. Grigsby, A.~E. Frias, V.~C.~C. Carlson, and C.~D.
  Kroenke, ``Folding, but not surface area expansion, is associated with
  cellular morphological maturation in the fetal cerebral cortex,''
  \emph{Journal of Neuroscience}, vol.~37, no.~8, pp. 1971--1983, 2017.

\bibitem{llinares2019deconstructing}
C.~Llinares-Benadero and V.~Borrell, ``Deconstructing cortical folding:
  genetic, cellular and mechanical determinants,'' \emph{Nature Reviews
  Neuroscience}, vol.~20, no.~3, pp. 161--176, 2019.

\bibitem{rana2019subplate}
S.~Rana, R.~Shishegar, S.~Quezada, L.~Johnston, D.~W. Walker, and M.~Tolcos,
  ``The subplate: a potential driver of cortical folding?'' \emph{Cerebral
  Cortex}, vol.~29, no.~11, pp. 4697--4708, 2019.

\bibitem{clouchoux2013delayed}
C.~Clouchoux, A.~Du~Plessis, M.~Bouyssi-Kobar, W.~Tworetzky, D.~McElhinney,
  D.~Brown \emph{et~al.}, ``Delayed cortical development in fetuses with
  complex congenital heart disease,'' \emph{Cerebral cortex}, vol.~23, no.~12,
  pp. 2932--2943, 2013.

\bibitem{im2017quantitative}
K.~Im, A.~Guimaraes, Y.~Kim, E.~Cottrill, B.~Gagoski, C.~Rollins \emph{et~al.},
  ``Quantitative folding pattern analysis of early primary sulci in human
  fetuses with brain abnormalities,'' \emph{American Journal of
  Neuroradiology}, vol.~38, no.~7, pp. 1449--1455, 2017.

\bibitem{tarui2018disorganized}
T.~Tarui, N.~Madan, N.~Farhat, R.~Kitano, A.~Ceren~Tanritanir, G.~Graham
  \emph{et~al.}, ``Disorganized patterns of sulcal position in fetal brains
  with agenesis of corpus callosum,'' \emph{Cerebral Cortex}, vol.~28, no.~9,
  pp. 3192--3203, 2018.

\bibitem{benkarim2018cortical}
O.~M. Benkarim, N.~Hahner, G.~Piella, E.~Gratacos, M.~A.~G. Ballester,
  E.~Eixarch \emph{et~al.}, ``Cortical folding alterations in fetuses with
  isolated non-severe ventriculomegaly,'' \emph{NeuroImage: Clinical}, vol.~18,
  pp. 103--114, 2018.

\bibitem{ortinau2019early}
C.~M. Ortinau, C.~K. Rollins, A.~Gholipour, H.~J. Yun, M.~Marshall, B.~Gagoski
  \emph{et~al.}, ``Early-emerging sulcal patterns are atypical in fetuses with
  congenital heart disease,'' \emph{Cerebral Cortex}, vol.~29, no.~8, pp.
  3605--3616, 2019.

\bibitem{gholipour2010robust}
A.~Gholipour, J.~A. Estroff, and S.~K. Warfield, ``Robust super-resolution
  volume reconstruction from slice acquisitions: application to fetal brain
  {MRI},'' \emph{IEEE transactions on medical imaging}, vol.~29, no.~10, pp.
  1739--1758, 2010.

\bibitem{studholme2011mapping}
C.~Studholme, ``Mapping fetal brain development in utero using magnetic
  resonance imaging: the big bang of brain mapping,'' \emph{Annual review of
  biomedical engineering}, vol.~13, pp. 345--368, 2011.

\bibitem{kuklisova2012reconstruction}
M.~Kuklisova-Murgasova, G.~Quaghebeur, M.~A. Rutherford, J.~V. Hajnal, and
  J.~A. Schnabel, ``Reconstruction of fetal brain {MRI} with intensity matching
  and complete outlier removal,'' \emph{Medical image analysis}, vol.~16,
  no.~8, pp. 1550--1564, 2012.

\bibitem{kainz2015fast}
B.~Kainz, M.~Steinberger, W.~Wein, M.~Kuklisova-Murgasova, C.~Malamateniou,
  K.~Keraudren \emph{et~al.}, ``Fast volume reconstruction from motion
  corrupted stacks of 2d slices,'' \emph{IEEE transactions on medical imaging},
  vol.~34, no.~9, pp. 1901--1913, 2015.

\bibitem{alansary2017pvr}
A.~Alansary, M.~Rajchl, S.~G. McDonagh, M.~Murgasova, M.~Damodaram, D.~F. Lloyd
  \emph{et~al.}, ``Pvr: Patch-to-volume reconstruction for large area motion
  correction of fetal {MRI},'' \emph{IEEE transactions on medical imaging},
  vol.~36, no.~10, pp. 2031--2044, 2017.

\bibitem{ebner2020automated}
M.~Ebner, G.~Wang, W.~Li, M.~Aertsen, P.~A. Patel, R.~Aughwane \emph{et~al.},
  ``An automated framework for localization, segmentation and super-resolution
  reconstruction of fetal brain {MRI},'' \emph{NeuroImage}, vol. 206, p.
  116324, 2020.

\bibitem{gholipour2017normative}
A.~Gholipour, C.~K. Rollins, C.~Velasco-Annis, A.~Ouaalam, A.~Akhondi-Asl,
  O.~Afacan \emph{et~al.}, ``A normative spatiotemporal {MRI} atlas of the
  fetal brain for automatic segmentation and analysis of early brain growth,''
  \emph{Scientific reports}, vol.~7, no.~1, p. 476, 2017.

\bibitem{habas2010atlas}
P.~A. Habas, K.~Kim, F.~Rousseau, O.~A. Glenn, A.~J. Barkovich, and
  C.~Studholme, ``Atlas-based segmentation of developing tissues in the human
  brain with quantitative validation in young fetuses,'' \emph{Human brain
  mapping}, vol.~31, no.~9, pp. 1348--1358, 2010.

\bibitem{makropoulos2018review}
A.~Makropoulos, S.~J. Counsell, and D.~Rueckert, ``A review on automatic fetal
  and neonatal brain {MRI} segmentation,'' \emph{NeuroImage}, vol. 170, pp.
  231--248, 2018.

\bibitem{xue2007automatic}
H.~Xue, L.~Srinivasan, S.~Jiang, M.~Rutherford, A.~D. Edwards, D.~Rueckert
  \emph{et~al.}, ``Automatic segmentation and reconstruction of the cortex from
  neonatal {MRI},'' \emph{Neuroimage}, vol.~38, no.~3, pp. 461--477, 2007.

\bibitem{weisenfeld2009automatic}
N.~I. Weisenfeld and S.~K. Warfield, ``Automatic segmentation of newborn brain
  {MRI},'' \emph{Neuroimage}, vol.~47, no.~2, pp. 564--572, 2009.

\bibitem{habas2008atlas}
P.~A. Habas, K.~Kim, F.~Rousseau, O.~A. Glenn, A.~J. Barkovich, and
  C.~Studholme, ``Atlas-based segmentation of the germinal matrix from in utero
  clinical {MRI} of the fetal brain,'' in \emph{International Conference on
  Medical Image Computing and Computer-Assisted Intervention}.\hskip 1em plus
  0.5em minus 0.4em\relax Springer, 2008, pp. 351--358.

\bibitem{habas2009statistical}
P.~A. Habas, K.~Kim, D.~Chandramohan, F.~Rousseau, O.~A. Glenn, and
  C.~Studholme, ``Statistical model of laminar structure for atlas-based
  segmentation of the fetal brain from in utero mr images,'' in \emph{Medical
  Imaging 2009: Image Processing}, vol. 7259.\hskip 1em plus 0.5em minus
  0.4em\relax International Society for Optics and Photonics, 2009, p. 725917.

\bibitem{habas2010spatiotemporal}
P.~A. Habas, K.~Kim, J.~M. Corbett-Detig, F.~Rousseau, O.~A. Glenn, A.~J.
  Barkovich \emph{et~al.}, ``A spatiotemporal atlas of mr intensity, tissue
  probability and shape of the fetal brain with application to segmentation,''
  \emph{Neuroimage}, vol.~53, no.~2, pp. 460--470, 2010.

\bibitem{serag2012multi}
A.~Serag, V.~Kyriakopoulou, M.~Rutherford, A.~Edwards, J.~Hajnal, P.~Aljabar
  \emph{et~al.}, ``A multi-channel 4d probabilistic atlas of the developing
  brain: application to fetuses and neonates,'' \emph{Annals of the BMVA}, vol.
  2012, no.~3, pp. 1--14, 2012.

\bibitem{gholipour2012multi}
A.~Gholipour, A.~Akhondi-Asl, J.~A. Estroff, and S.~K. Warfield, ``Multi-atlas
  multi-shape segmentation of fetal brain {MRI} for volumetric and morphometric
  analysis of ventriculomegaly,'' \emph{NeuroImage}, vol.~60, no.~3, pp.
  1819--1831, 2012.

\bibitem{gholipour2014construction}
A.~Gholipour, C.~Limperopoulos, S.~Clancy, C.~Clouchoux, A.~Akhondi-Asl, J.~A.
  Estroff \emph{et~al.}, ``Construction of a deformable spatiotemporal {MRI}
  atlas of the fetal brain: evaluation of similarity metrics and deformation
  models,'' in \emph{International Conference on Medical Image Computing and
  Computer-Assisted Intervention}.\hskip 1em plus 0.5em minus 0.4em\relax
  Springer, 2014, pp. 292--299.

\bibitem{khalili2019automatic}
N.~Khalili, N.~Lessmann, E.~Turk, N.~Claessens, R.~de~Heus, T.~Kolk
  \emph{et~al.}, ``Automatic brain tissue segmentation in fetal {MRI} using
  convolutional neural networks,'' \emph{Magnetic resonance imaging}, vol.~64,
  pp. 77--89, 2019.

\bibitem{cciccek20163d}
{\"O}.~{\c{C}}i{\c{c}}ek, A.~Abdulkadir, S.~S. Lienkamp, T.~Brox, and
  O.~Ronneberger, ``3d u-net: learning dense volumetric segmentation from
  sparse annotation,'' in \emph{International conference on medical image
  computing and computer-assisted intervention}.\hskip 1em plus 0.5em minus
  0.4em\relax Springer, 2016, pp. 424--432.

\bibitem{Hong2019FetalCP}
J.~Hong, H.~J. Yun, G.~Park, Y.-H. Park, J.-M. Lee, and K.~Im, ``Fetal cortical
  plate segmentation using 2d u-net with plane aggregation.''

\bibitem{oktay2018attention}
O.~Oktay, J.~Schlemper, L.~L. Folgoc, M.~Lee, M.~Heinrich, K.~Misawa
  \emph{et~al.}, ``Attention u-net: Learning where to look for the pancreas,''
  \emph{arXiv preprint arXiv:1804.03999}, 2018.

\bibitem{roy2018recalibrating}
A.~G. Roy, N.~Navab, and C.~Wachinger, ``Recalibrating fully convolutional
  networks with spatial and channel “squeeze and excitation” blocks,''
  \emph{IEEE transactions on medical imaging}, vol.~38, no.~2, pp. 540--549,
  2018.

\bibitem{salehi2018real}
S.~S.~M. Salehi, S.~R. Hashemi, C.~Velasco-Annis, A.~Ouaalam, J.~A. Estroff,
  D.~Erdogmus \emph{et~al.}, ``Real-time automatic fetal brain extraction in
  fetal {MRI} by deep learning,'' in \emph{2018 IEEE 15th International
  Symposium on Biomedical Imaging (ISBI 2018)}.\hskip 1em plus 0.5em minus
  0.4em\relax IEEE, 2018, pp. 720--724.

\bibitem{salehi2017auto}
S.~S.~M. Salehi, D.~Erdogmus, and A.~Gholipour, ``Auto-context convolutional
  neural network (auto-net) for brain extraction in magnetic resonance
  imaging,'' \emph{IEEE transactions on medical imaging}, vol.~36, no.~11, pp.
  2319--2330, 2017.

\bibitem{yushkevich2019user}
P.~A. Yushkevich, A.~Pashchinskiy, I.~Oguz, S.~Mohan, J.~E. Schmitt, J.~M.
  Stein \emph{et~al.}, ``User-guided segmentation of multi-modality medical
  imaging datasets with itk-snap,'' \emph{Neuroinformatics}, vol.~17, no.~1,
  pp. 83--102, 2019.

\bibitem{he2016deep}
K.~He, X.~Zhang, S.~Ren, and J.~Sun, ``Deep residual learning for image
  recognition,'' in \emph{Proceedings of the IEEE conference on computer vision
  and pattern recognition}, 2016, pp. 770--778.

\bibitem{ioffe2015batch}
S.~Ioffe and C.~Szegedy, ``Batch normalization: Accelerating deep network
  training by reducing internal covariate shift,'' \emph{arXiv preprint
  arXiv:1502.03167}, 2015.

\bibitem{he2015delving}
K.~He, X.~Zhang, S.~Ren, and J.~Sun, ``Delving deep into rectifiers: Surpassing
  human-level performance on imagenet classification,'' in \emph{Proceedings of
  the IEEE international conference on computer vision}, 2015, pp. 1026--1034.

\bibitem{xie2015holistically}
S.~Xie and Z.~Tu, ``Holistically-nested edge detection,'' in \emph{Proceedings
  of the IEEE Int. conference on computer vision}, 2015, pp. 1395--1403.

\bibitem{xie2017aggregated}
S.~Xie, R.~Girshick, P.~Doll{\'a}r, Z.~Tu, and K.~He, ``Aggregated residual
  transformations for deep neural networks,'' in \emph{Proc. of the IEEE
  conference on computer vision and pattern recognition}, 2017, pp. 1492--1500.

\bibitem{paszke2019pytorch}
A.~Paszke, S.~Gross, F.~Massa, A.~Lerer, J.~Bradbury, G.~Chanan \emph{et~al.},
  ``Pytorch: An imperative style, high-performance deep learning library,'' in
  \emph{Advances in Neural Information Processing Systems}, 2019, pp.
  8024--8035.

\bibitem{kingma2014adam}
D.~P. Kingma and J.~Ba, ``Adam: A method for stochastic optimization,''
  \emph{arXiv preprint arXiv:1412.6980}, 2014.

\bibitem{ronneberger2015u}
O.~Ronneberger, P.~Fischer, and T.~Brox, ``U-net: Convolutional networks for
  biomedical image segmentation,'' in \emph{International Conference on Medical
  image computing and computer-assisted intervention}.\hskip 1em plus 0.5em
  minus 0.4em\relax Springer, 2015, pp. 234--241.

\bibitem{milletari2016v}
F.~Milletari, N.~Navab, and S.-A. Ahmadi, ``V-net: Fully convolutional neural
  networks for volumetric medical image segmentation,'' in \emph{2016 Fourth
  International Conference on 3D Vision (3DV)}.\hskip 1em plus 0.5em minus
  0.4em\relax IEEE, 2016, pp. 565--571.

\bibitem{salehi2017tversky}
S.~S.~M. Salehi, D.~Erdogmus, and A.~Gholipour, ``Tversky loss function for
  image segmentation using 3d fully convolutional deep networks,'' in
  \emph{International Workshop on Machine Learning in Medical Imaging}.\hskip
  1em plus 0.5em minus 0.4em\relax Springer, 2017, pp. 379--387.

\bibitem{du2020medical}
G.~Du, X.~Cao, J.~Liang, X.~Chen, and Y.~Zhan, ``Medical image segmentation
  based on u-net: A review,'' \emph{Journal of Imaging Science and Technology},
  2020.

\bibitem{minaee2020image}
S.~Minaee, Y.~Boykov, F.~Porikli, A.~Plaza, N.~Kehtarnavaz, and D.~Terzopoulos,
  ``Image segmentation using deep learning: A survey,'' \emph{arXiv preprint
  arXiv:2001.05566}, 2020.

\bibitem{jegou2017one}
S.~J{\'e}gou, M.~Drozdzal, D.~Vazquez, A.~Romero, and Y.~Bengio, ``The one
  hundred layers tiramisu: Fully convolutional densenets for semantic
  segmentation,'' in \emph{Proceedings of the IEEE conference on computer
  vision and pattern recognition workshops}, 2017, pp. 11--19.

\bibitem{wang2019automatic}
G.~Wang, J.~Shapey, W.~Li, R.~Dorent, A.~Demitriadis, S.~Bisdas \emph{et~al.},
  ``Automatic segmentation of vestibular schwannoma from t2-weighted mri by
  deep spatial attention with hardness-weighted loss,'' in \emph{International
  Conference on Medical Image Computing and Computer-Assisted
  Intervention}.\hskip 1em plus 0.5em minus 0.4em\relax Springer, 2019, pp.
  264--272.

\bibitem{avants2011reproducible}
B.~B. Avants, N.~J. Tustison, G.~Song, P.~A. Cook, A.~Klein, and J.~C. Gee, ``A
  reproducible evaluation of ants similarity metric performance in brain image
  registration,'' \emph{Neuroimage}, vol.~54, no.~3, pp. 2033--2044, 2011.

\bibitem{akhondi2013simultaneous}
A.~Akhondi-Asl and S.~K. Warfield, ``Simultaneous truth and performance level
  estimation through fusion of probabilistic segmentations,'' \emph{IEEE
  transactions on medical imaging}, vol.~32, no.~10, pp. 1840--1852, 2013.

\bibitem{chang2009performance}
H.-H. Chang, A.~H. Zhuang, D.~J. Valentino, and W.-C. Chu, ``Performance
  measure characterization for evaluating neuroimage segmentation algorithms,''
  \emph{Neuroimage}, vol.~47, no.~1, pp. 122--135, 2009.

\bibitem{mendrik2015mrbrains}
A.~M. Mendrik, K.~L. Vincken, H.~J. Kuijf, M.~Breeuwer, W.~H. Bouvy,
  J.~De~Bresser \emph{et~al.}, ``Mrbrains challenge: online evaluation
  framework for brain image segmentation in 3t {MRI} scans,''
  \emph{Computational intelligence and neuroscience}, vol. 2015, p.~1, 2015.

\bibitem{yeghiazaryan2015overview}
V.~Yeghiazaryan and I.~Voiculescu, ``An overview of current evaluation methods
  used in medical image segmentation,'' \emph{Tech. Rep. RR-15-08, Department
  of Computer Science}, 2015.

\bibitem{hesamian2019deep}
M.~H. Hesamian, W.~Jia, X.~He, and P.~Kennedy, ``Deep learning techniques for
  medical image segmentation: Achievements and challenges,'' \emph{Journal of
  digital imaging}, vol.~32, no.~4, pp. 582--596, 2019.

\bibitem{hu2018squeeze}
J.~Hu, L.~Shen, and G.~Sun, ``Squeeze-and-excitation networks,'' in
  \emph{Proceedings of the IEEE conference on computer vision and pattern
  recognition}, 2018, pp. 7132--7141.

\bibitem{chen2016attention}
L.-C. Chen, Y.~Yang, J.~Wang, W.~Xu, and A.~L. Yuille, ``Attention to scale:
  Scale-aware semantic image segmentation,'' in \emph{Proceedings of the IEEE
  conference on computer vision and pattern recognition}, 2016, pp. 3640--3649.

\bibitem{woo2018cbam}
S.~Woo, J.~Park, J.-Y. Lee, and I.~So~Kweon, ``Cbam: Convolutional block
  attention module,'' in \emph{Proceedings of the European Conference on
  Computer Vision (ECCV)}, 2018, pp. 3--19.

\bibitem{zhang2018self}
H.~Zhang, I.~Goodfellow, D.~Metaxas, and A.~Odena, ``Self-attention generative
  adversarial networks,'' \emph{arXiv preprint arXiv:1805.08318}, 2018.

\bibitem{szegedy2015going}
C.~Szegedy, W.~Liu, Y.~Jia, P.~Sermanet, S.~Reed, D.~Anguelov \emph{et~al.},
  ``Going deeper with convolutions,'' in \emph{Proceedings of the IEEE
  conference on computer vision and pattern recognition}, 2015, pp. 1--9.

\bibitem{tan2019mixconv}
M.~Tan and Q.~V. Le, ``Mixconv: Mixed depthwise convolutional kernels,''
  \emph{CoRR, abs/1907.09595}, 2019.

\bibitem{zhao2017pyramid}
H.~Zhao, J.~Shi, X.~Qi, X.~Wang, and J.~Jia, ``Pyramid scene parsing network,''
  in \emph{Proceedings of the IEEE conference on computer vision and pattern
  recognition}, 2017, pp. 2881--2890.

\end{thebibliography}
		
	\clearpage
	\includepdfmerge{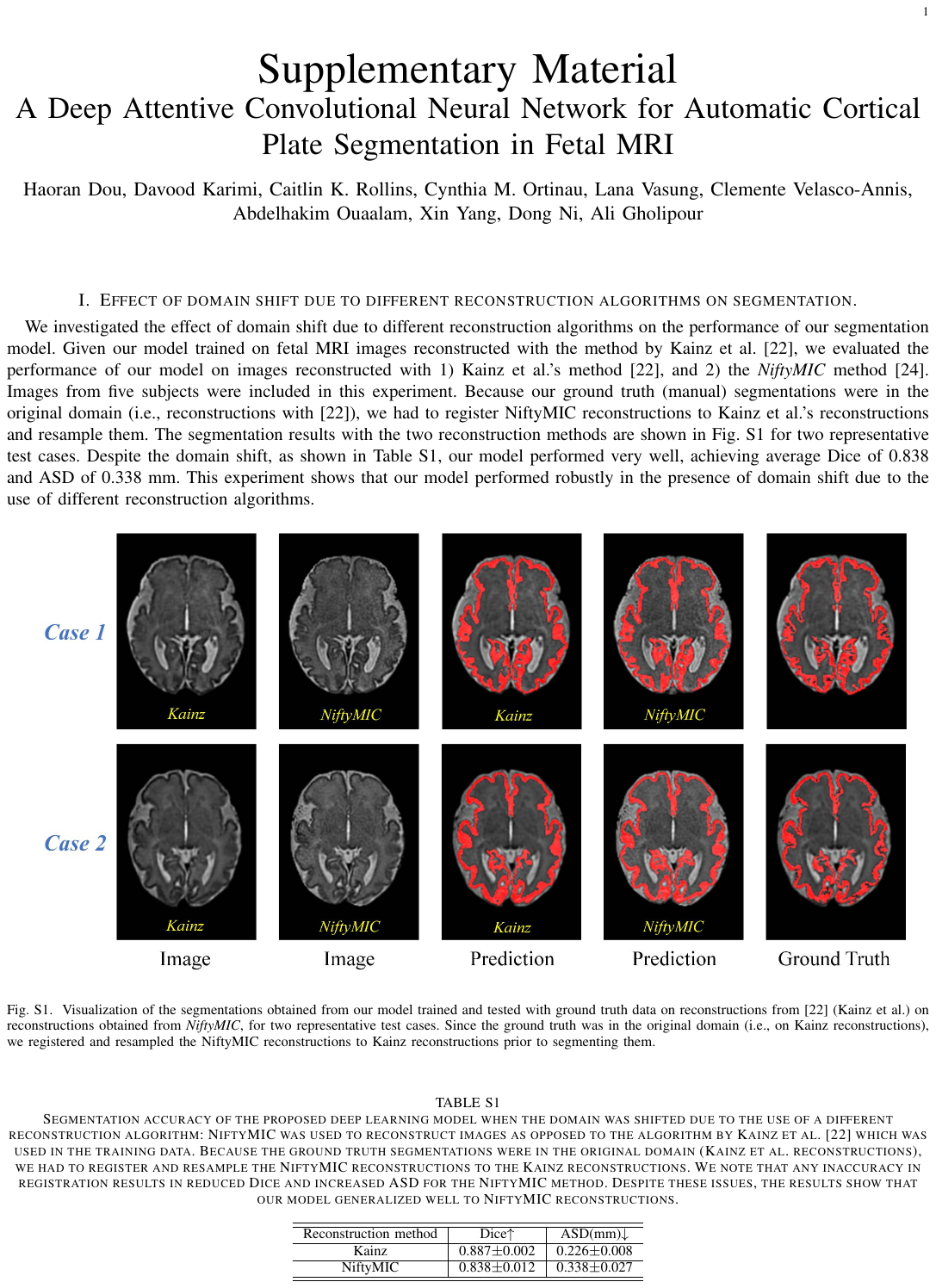, 1}
	
	\ifCLASSOPTIONcaptionsoff
	\newpage
	\fi

\end{document}